\documentclass[letterpaper, 10 pt, conference]{ieeeconf} 
\IEEEoverridecommandlockouts
\usepackage{amsmath,amsfonts}
\usepackage{algorithmic}
\usepackage{algorithm}
\usepackage{array}
\usepackage[caption=false,font=normalsize,labelfont=sf,textfont=sf]{subfig}
\usepackage{textcomp}
\usepackage{stfloats}
\usepackage{url}
\usepackage{verbatim}
\usepackage{graphicx}
\usepackage{cite}
\usepackage{multirow}
\usepackage{caption}
\usepackage{siunitx}

\usepackage{xcolor}

\makeatletter
\let\NAT@parse\undefined
\makeatother

\usepackage{hyperref}
 \usepackage{orcidlink}
\hypersetup{
    colorlinks=true,
    linkcolor=blue,   
    citecolor=blue,   
    urlcolor=blue    
}

\let\oldcite=\cite
\renewcommand{\cite}[1]{\textcolor{blue}{\oldcite{#1}}}

\title{\LARGE \bf
DTP-Attack: A decision-based black-box adversarial attack on trajectory prediction}
\author{Jiaxiang Li$^{1}$, Jun Yan$^{1}$, Daniel Watzenig$^2$ and Huilin Yin$^{1}$
\thanks{$^{1}$Jiaxiang Li, Jun Yan and Huilin Yin (e-mail: zmbdsilver@gmail.com, yanjun@tongji.edu.cn, yinhuilin@tongji.edu.cn) are with the College of Electronics and Information Engineering, Tongji University, Shanghai 201804, China. (Corresponding author: Huilin Yin.)}
\thanks{$^{2}$Daniel Watzenig (e-mail: daniel.watzenig@tugraz.at) is with the Graz University of Technology and the Virtual Vehicle Research, Graz 8010, Austria.}
}

\begin{document}

\maketitle

\thispagestyle{empty}
\pagestyle{empty}

\begin{abstract}
Trajectory prediction systems are critical for autonomous vehicle safety, yet remain vulnerable to adversarial attacks that can cause catastrophic traffic behavior misinterpretations. Existing attack methods require white-box access with gradient information and rely on rigid physical constraints, limiting real-world applicability.
We propose DTP-Attack, a decision-based black-box adversarial attack framework tailored for trajectory prediction systems. Our method operates exclusively on binary decision outputs without requiring model internals or gradients, making it practical for real-world scenarios. DTP-Attack employs a novel boundary walking algorithm that navigates adversarial regions without fixed constraints, naturally maintaining trajectory realism through proximity preservation. Unlike existing approaches, our method supports both intention misclassification attacks and prediction accuracy degradation.
Extensive evaluation on nuScenes and Apolloscape datasets across state-of-the-art models including Trajectron++ and Grip++ demonstrates superior performance. DTP-Attack achieves $41-81\%$ attack success rates for intention misclassification attacks that manipulate perceived driving maneuvers with perturbations below \SI{0.45}{\meter}, and increases prediction errors by $1.9- 4.2\times $ for accuracy degradation. Our method consistently outperforms existing black-box approaches while maintaining high controllability and reliability across diverse scenarios. These results reveal fundamental vulnerabilities in current trajectory prediction systems, highlighting urgent needs for robust defenses in safety-critical autonomous driving applications. 
Our code is available at the repository: \url{https://github.com/eclipse-bot/DTP-Attack}.
\end{abstract}

\section{Introduction}
Autonomous vehicles (AVs) rely critically on trajectory prediction systems\cite{salzmann2020trajectron++, li2019grip} to forecast the future movements of surrounding traffic participants. These deep learning-based models enable vehicles to anticipate potential hazards and make safe navigation decisions, forming a cornerstone of modern autonomous driving technology. However, recent advances in adversarial machine learning\cite{he2023adversarial} reveal that neural networks can be manipulated through subtle input perturbations, causing incorrect predictions while appearing to function normally. In autonomous driving, such vulnerabilities pose severe safety risks, as a compromised trajectory prediction system could cause catastrophic misinterpretations of nearby traffic behavior.

\begin{figure}[!t]
\centering
\includegraphics[width=0.99\linewidth]{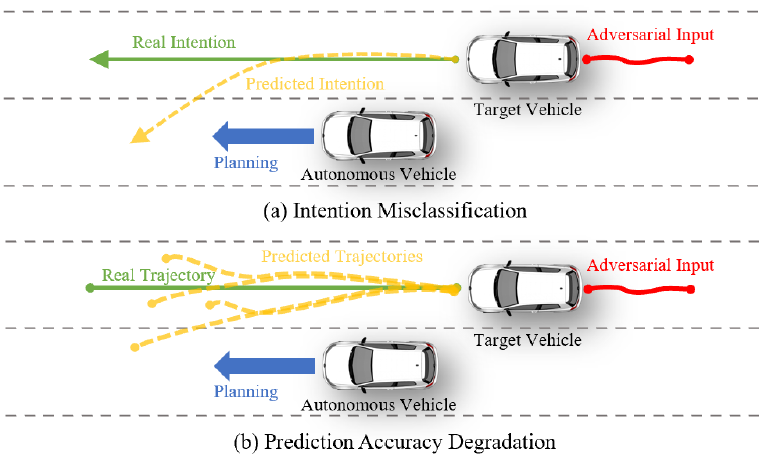}
\caption{Two types of adversarial attacks on trajectory prediction. (a) Intention misclassification. (b) Prediction accuracy degradation.}
\label{fig:goal}
\end{figure}

The threat model for trajectory prediction attacks encompasses two primary objectives, as illustrated in Fig. \ref{fig:goal}. Intention misclassification attacks\cite{choi2021drogon} aim to manipulate the perceived driving maneuvers of target vehicles, causing a vehicle traveling straight to appear as though it will execute a lane change or turn. Prediction accuracy degradation attacks systematically increase standard trajectory prediction error metrics, corrupting the system's ability to accurately forecast future positions and potentially compromising downstream planning algorithms.

\begin{figure}[!t]
\centering
\includegraphics[width=0.99\linewidth]{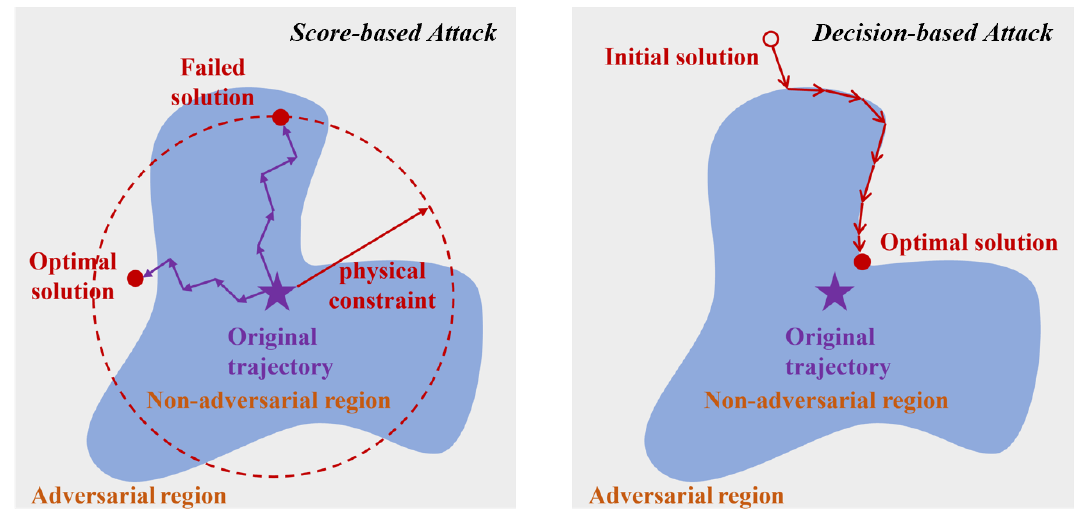}
\caption{Conceptual comparison of score-based (left) vs. decision-based (right) adversarial attack optimization landscapes.}
\label{fig:detail}
\end{figure}

Compared to traditional image classification domains\cite{machado2021adversarial, ZHU2024128512}, adversarial attacks on trajectory prediction systems present distinct challenges. First, trajectory data consists of spatiotemporal sequences that must satisfy kinematic constraints, since arbitrary perturbations can easily violate realistic vehicle dynamics\cite{zhong2023guided}. Second, trajectory prediction models typically output continuous coordinate sequences rather than discrete class probabilities, complicating the definition of attack objectives. Third, real-world threat scenarios often involve black-box access\cite{kumar2020black, ilyas2018black} where attackers cannot observe model internals or gradients.

As illustrated in Fig. \ref{fig:detail}, existing adversarial attack methods\cite{zhang2022adversarial,cao2022advdo,10802234,tan2023targeted} for trajectory prediction employ score-based optimization paradigms that exhibit fundamental structural limitations. Current approaches universally rely on gradient information or detailed model access while imposing artificially specified physical constraints that inadequately capture real-world vehicle dynamics. This score-based formulation creates the discontinuous search landscapes depicted in the figure, where rigid constraint boundaries systematically trap optimization algorithms in local optima\cite{xu2023constrained}, preventing convergence to effective solutions and generating unrealistic trajectories. Moreover, score-based methods are inherently restricted to prediction accuracy degradation objectives, fundamentally precluding adaptation to intention misclassification attacks and significantly limiting their applicability in realistic threat scenarios.


We propose DTP-Attack, a decision-based black-box adversarial attack framework that addresses these limitations through a fundamentally different optimization paradigm. Our approach operates exclusively on binary feedback from adversarial criteria functions, eliminating the requirement for gradient information or model internals. As demonstrated in Fig. \ref{fig:detail}, this decision-based formulation creates smooth adversarial regions that enable effective boundary walking optimization. Rather than imposing explicit physical constraints, our method maintains trajectory realism through proximity preservation, allowing unconstrained exploration of the adversarial space while naturally satisfying kinematic feasibility. The proposed boundary walking algorithm iteratively refines adversarial trajectories through orthogonal and forward steps along the decision boundary, consistently converging to near-optimal solutions. 

Our experimental evaluation across multiple state-of-the-art models and datasets demonstrates superior attack effectiveness compared to existing black-box baselines. DTP-Attack achieves attack success rates ranging from $41\%$ to $81\%$ for intention misclassification while maintaining perturbation magnitudes below \SI{0.45}{\meter}. For accuracy degradation objectives, our method increases prediction errors by factors of $1.9\times$ to $4.2\times$, significantly compromising system reliability through minimal trajectory modifications.

Our main contributions are threefold: First, we propose a decision-based black-box attack method for trajectory prediction that operates without requiring model internals or gradient information, making it applicable to real-world scenarios. Second, we develop a constraint-free optimization approach that naturally maintains trajectory realism while allowing attackers to control attack effects according to their objectives. Third, we provide comprehensive experimental validation showing that our method consistently outperforms existing approaches while maintaining attack stealthiness and reliability across diverse model architectures and datasets.
\section{Related Work}
\subsection{Trajectory Prediction in Autonomous Driving}
Neural trajectory prediction systems enable autonomous vehicles to forecast future movements of surrounding agents based on historical observations and environmental context. Leading approaches include Trajectron++\cite{salzmann2020trajectron++}, which models driving scenarios as directed graphs to capture agent interactions and incorporates vehicle dynamics for realistic predictions, and Grip++\cite{li2019grip}, which uses graph convolutions with temporal modeling. While these methods achieve strong performance on standard benchmarks like nuScenes\cite{caesar2020nuscenes} and Apolloscape\cite{huang2018apolloscape}, they inherit the vulnerability of deep neural networks to adversarial perturbations.
\subsection{Adversarial Attacks on Trajectory Prediction}
Adversarial attacks on trajectory prediction present unique challenges compared to traditional domains like image classification, as trajectory data must satisfy physical constraints while maintaining temporal coherence. Existing methods can be categorized based on the level of target model access required.

White-box attacks leverage complete model knowledge but face significant practical limitations. Zhang et al.\cite{zhang2022adversarial} applied Projected Gradient Descent (PGD) to spatial coordinates, though this produced unrealistic trajectories. Cao et al.\cite{cao2022advdo} improved physical feasibility by perturbing control signals and recovering trajectories through dynamic models. Tan et al.\cite{tan2023targeted} extended this work to targeted attacks that deceive models into predicting user-specified outcomes. However, these methods require proprietary model internals and rely on manually specified constraints that fail to capture real-world vehicle dynamics complexity.

Black-box attacks offer greater practical relevance but suffer from fundamental limitations. Zhang et al.\cite{zhang2022adversarial} adapted Particle Swarm Optimization (PSO) for trajectory attacks, but PSO's design for unconstrained optimization poorly suits the nonlinear constrained problems inherent in realistic trajectory generation. Critically, existing approaches rely on score-based optimization requiring continuous confidence scores or gradient information. This creates discontinuous search landscapes that trap algorithms in local optima and limit convergence. Moreover, current methods focus primarily on prediction accuracy degradation and lack flexibility for diverse attack objectives such as intention misclassification.

Our decision-based framework addresses these limitations by operating exclusively on binary decision outputs without requiring gradient information or manually specified constraints, enabling more effective optimization across diverse attack objectives.

\begin{figure*}[!t]
\centering
\includegraphics[width=0.92\textwidth]{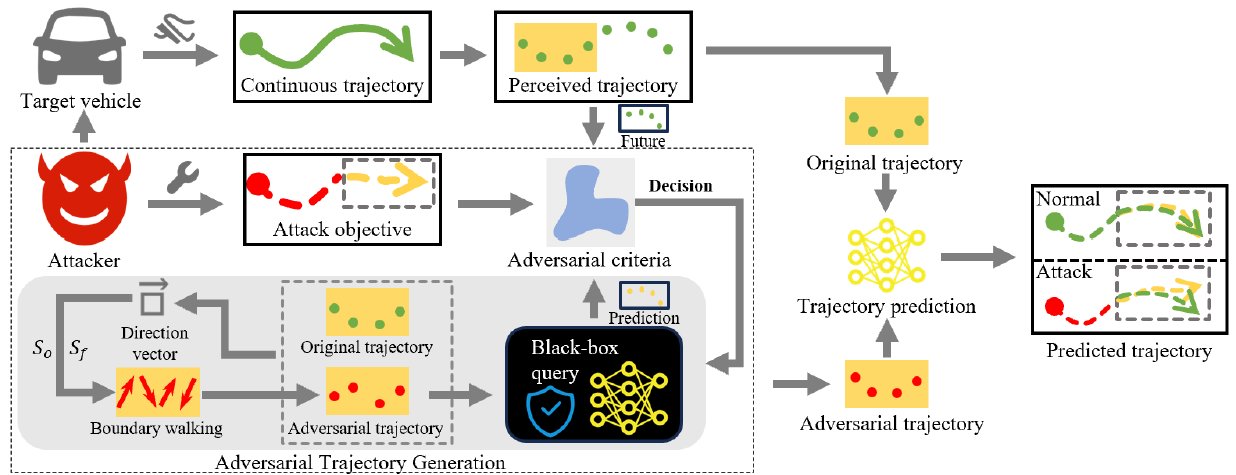}
\caption{A decision-based black-box adversarial attack on trajectory prediction (DTP-Attack) methodology overview.}
\label{fig:method}
\end{figure*}

\section{Problem Formulation}
\subsection{Trajectory Prediction Formulation}

We address the trajectory prediction problem, which operates at fixed time intervals to forecast future movement patterns for multiple agents within a given scenario. The task requires predicting probability distributions over possible future trajectories for $N$ agents based on their historical states and contextual information.

We represent each agent's state as a $D$-dimensional vector $s\in \mathbb{R}^{D}$, which captures spatial coordinates along with additional attributes such as semantic class, agent dimensions, and heading direction. Our analysis concentrates on spatial coordinates $p \in \mathbb{R}^{2}$, representing each agent's position in two-dimensional space.

At any given time $t$, the prediction system receives historical state information spanning $L_{I}$ previous timesteps. We denote this input history as $\mathbf{H}_{t}$, $\mathbf{H}_{t}=s_{t-L_{I}+1:t}^{1,...,N} \in \mathbb{R}^{L_{I}\times N\times D}$, where the tensor dimensions represent time, agents, and state features respectively. The system also incorporates additional contextual information $\mathbf{I}_{t}$, which may include map data, traffic signals, or other environmental factors.

The system estimates probability distributions over future trajectories spanning the subsequent $L_{O}$ timesteps. We represent predicted trajectory coordinates as $\mathbf{P}_{t}=p_{t+1:t+L_{O}}^{1,...,N}\in \mathbb{R}^{L_{O}\times N\times 2}$ and ground truth future states as $\mathbf{F}_{t}=s_{t+1:t+L_{O}}^{1,...,N}\in \mathbb{R}^{L_{O}\times N\times D}$. The trajectory prediction model functions as $\mathbf{P}_t = \Phi(\mathbf{H}_t, \mathbf{I}_t)$, mapping historical states and contextual information to predicted trajectory distributions.

\subsection{Adversarial Attack Formulation}

Our attack operates in real-world autonomous driving environments by exploiting how AVs predict the future movements of surrounding traffic. In this scenario, an attacker uses a vehicle to execute specific movement patterns designed to mislead the AV's trajectory prediction system.

The approach works as follows: A vehicle near the AV is selected as the target vehicle with an original trajectory of $X_t \in \mathbf{H}_{t}$. The attacker drives this vehicle following a calculated movement pattern that, when processed by the AV's perception system, creates an adversarial trajectory $X_t^*$ derived from but subtly different than $X_t$. These small deviations are precisely engineered using our DTP-Attack method to remain within physical feasibility constraints while causing specific prediction errors.

We examine black-box attack scenarios where attackers possess no internal knowledge of the AV's trajectory prediction model architecture or parameters. Attackers can only observe system behavior through input-output relationships, making this approach more applicable to real-world conditions than attacks requiring detailed model access. This constraint increases practical relevance while presenting greater technical challenges for attack generation.
\section{Adversarial Trajectory Generation}
As shown in Fig. \ref{fig:method}, DTP-Attack transforms the attack objective into binary adversarial criteria that partition trajectory space into adversarial and non-adversarial regions. The framework operates through iterative black-box queries to the target model, employing boundary walking optimization with orthogonal and forward steps to navigate the adversarial region. This process systematically converges to minimal perturbations that achieve specified attack goals while preserving trajectory realism.
\subsection{Attack Objectives}
DTP-Attack targets two distinct categories of trajectory prediction failures: intention misclassification and prediction accuracy degradation. Each objective requires different manipulation strategies and evaluation metrics.

\textbf{Intention Misclassification}:
This objective aims to cause misclassification of intended maneuvers. For example, an attack might make a vehicle traveling straight appear to be turning left or right, or make a vehicle maintaining constant speed appear to be accelerating. To achieve this, predicted trajectories must be shifted in specific directions, such as leftward for simulating left turns, or forward along the longitudinal axis for simulating acceleration.

We design four directional metrics to calculate deviations: left turn, right turn (lateral direction), and forward, backward (longitudinal direction). The formula is defined as:
\begin{equation}\label{eqn-1} 
d_{int}\left (\mathbf{P}_{t}, \mathbf{F}_{t}, n\right )= \frac{1}{L_{O}}\sum_{i=t+1}^{t+L_{O}}\left ( p_{i}^{n}-s_{i}^{n} \right )^{T}\cdot \mathcal{G}  \left (s_{i+1}^{n},s_{i}^{n} \right) 
\end{equation}
where $n$ denotes the target vehicle ID, $p$ and $s$ represent predicted and ground-truth vehicle locations respectively, and $\mathcal{G}$ is a direction-specific unit vector generator. The longitudinal direction is approximated as $s_{i+1}^{n}-s_{i}^{n}$.

\textbf{Prediction Accuracy Degradation}:
This objective maximizes standard trajectory prediction error metrics. We focus on two commonly used measures: Average Displacement Error (ADE), which calculates the average root mean squared error between predicted and ground truth trajectories, and Final Displacement Error (FDE), which measures displacement error at the final predicted timestep.

The formulas are:
\begin{equation}\label{eqn-1} 
ADE=\frac{1}{L_{O}}\sum_{i=t+1}^{t+L_{O}}\left \| p_{i}^{n}-s_{i}^{n}  \right \| _{2}
\end{equation}
\begin{equation}\label{eqn-1} 
FDE=\left \| p_{t+L_{O} }^{n}- s_{t+L_{O} }^{n} \right \| _{2} 
\end{equation}

The combined attack objective metric is:
\begin{equation}\label{eqn-1} 
d_{err}\left ( \mathbf{P}_{t}, \mathbf{F}_{t}, n \right )=\mathbb{I} _{ADE}\cdot ADE+ \mathbb{I} _{FDE}\cdot FDE
\end{equation}
where $\mathbb{I} _{ADE}$ and $\mathbb{I} _{FDE}$ are indicator variables set to $1$ when targeting ADE and FDE respectively, and $0$ otherwise.

\subsection{Adversarial Criteria}
To establish a unified framework for adversarial detection, we define an adversarial criterion based on selectively activated attack objective functions. Let $D=\left \{ d_{int},d_{err}\right \} $ denote the set of available attack objective functions, with corresponding threshold set $\Theta =\left \{ \theta _{int},\theta _{err} \right \}$, where each threshold $\theta_{i}\in \Theta$ represents the minimum activation level required for the corresponding attack objective $d_{i}\in D$.

At each evaluation instance, only one attack objective function is selected from $D$. Let $d_{active}\in D$ denote the currently selected attack objective with its corresponding threshold $\theta_{active}\in \Theta$. The adversarial criteria function is expressed as:
\begin{equation}\label{eqn-1}
c(\mathbf{P}_{t},\mathbf{F}_{t},n)=\mathbb{I}\left ( d_{active}(\mathbf{P}_{t},\mathbf{F}_{t},n)> \theta_{active} \right ) 
\end{equation}
By incorporating the model relationship $\mathbf{P}_{t} = \Phi(X_{t}\cup \mathbf{H}_{t},\mathbf{I}_{t})$, we express the criteria in terms of input trajectory: $c(X_{t},\mathbf{F}_{t},n)$.

This adversarial criteria function serves as a binary decision output function that partitions the trajectory input space into adversarial and non-adversarial regions. The function outputs $1$ when the attack objective functions exceed their corresponding thresholds, and 0 otherwise.

\begin{algorithm}[!t]
\caption{Decision-based Attack Method}
\label{alg:dtp_attack}
\begin{algorithmic}[1]
\renewcommand{\algorithmicrequire}{\textbf{Input:}}
\renewcommand{\algorithmicensure}{\textbf{Output:}}
\REQUIRE Original trajectory $X_t$, historical states $\mathbf{H}_t$, contextual information $\mathbf{I}_t$, trajectory prediction model $\Phi(\cdot)$, adversarial criteria $c(\cdot)$, ground truth $\mathbf{F}_t$

\ENSURE Adversarial trajectory $X^*_t$ minimizing $\mathcal{D}(X_t, X^*_t)$

\STATE \textbf{Initialize:} $k \leftarrow 0$, $\delta \leftarrow 1.0$, $\epsilon \leftarrow 0.1$, $\text{max\_iter} \leftarrow 1000$
\STATE \textbf{Find initial adversarial point:} Sample $X^{*(0)}_t$ such that $c(\Phi(X^{*(0)}_t \cup \mathbf{H}_t, \mathbf{I}_t), \mathbf{F}_t) = 1$
\STATE \textbf{Forward initialization:} Perform initial forward step towards $X_t$

\WHILE{$k < \text{max\_iter}$ \AND $\epsilon > \text{tolerance}$}
    \REPEAT
        \STATE Orthogonal direction: $\mathbf{d}_{\perp} \leftarrow \perp(X_t - X^{*(k)}_t)$
        \STATE Sample orthogonal step: $S_o^{(k)} \leftarrow \delta \cdot \mathbf{d}_{\perp} \cdot \mathcal{N}(0,1)$
        \STATE $X_{\text{temp}} \leftarrow X^{*(k)}_t + S_o^{(k)}$
        \IF{$c(\Phi(X_{\text{temp}} \cup \mathbf{H}_t, \mathbf{I}_t), \mathbf{F}_t) = 1$}
            \STATE \textbf{break} 
        \ELSE
            \STATE $\delta \leftarrow \delta \times 0.95$ 
        \ENDIF
    \UNTIL{valid orthogonal step found}
    
    \REPEAT
        \STATE Forward direction: $\mathbf{d}_f \leftarrow \frac{X_t - (X^{*(k)}_t + S_o^{(k)})}{\|X_t - (X^{*(k)}_t + S_o^{(k)})\|}$
        \STATE Compute forward step: $S_f^{(k)} \leftarrow \epsilon \cdot \mathbf{d}_f$
        \STATE $X_{\text{candidate}} \leftarrow X^{*(k)}_t + S_o^{(k)} + S_f^{(k)}$
        \IF{$c(\Phi(X_{\text{candidate}} \cup \mathbf{H}_t, \mathbf{I}_t), \mathbf{F}_t) = 1$}
            \STATE \textbf{break} 
        \ELSE
            \STATE $\epsilon \leftarrow \epsilon \times 0.9$ 
        \ENDIF
    \UNTIL{valid forward step found \OR $\epsilon < 10^{-6}$}
    
    \STATE \textbf{Update:} $X^{*(k+1)}_t \leftarrow X^{*(k)}_t + S_o^{(k)} + S_f^{(k)}$
    \STATE $k \leftarrow k + 1$
    
    \IF{$\mathcal{D}(X_t, X^{*(k)}_t) - \mathcal{D}(X_t, X^{*(k-1)}_t) < 10^{-8}$}
        \STATE \textbf{break}
    \ENDIF
\ENDWHILE

\RETURN $X^*_t \leftarrow X^{*(k)}_t$
\end{algorithmic}
\end{algorithm}

\subsection{Adversarial Optimization Process}
Our approach fundamentally differs from existing score-based attack methods\cite{zhang2022adversarial,cao2022advdo,10802234}. Traditional score-based attacks utilize gradient information or stochastic exploration to generate adversarial trajectories through iterative optimization:
\begin{equation}\label{eqn-1}
\underset{X^{*}_{t} }{min}\ \zeta(X^{*}_{t}\in \mathbb{B}_{X,\gamma})+\mathcal{S}(\Phi (X^{*}_{t}\cup \mathbf{H}_{t},\mathbf{I}_{t}),\mathbf{F}_{t},n)
\end{equation}
Here, $\mathbb{B}_{X,\gamma}$ denotes a $\ell_{p}$ norm ball centered at $X_{t}$ with radius $\gamma$. The term $\zeta(.)$ represents a physical constraint function: $\zeta(u)=0$ if $X^{*}_{t}$ lies within ball $\mathbb{B}_{X,\gamma}$, otherwise $\zeta(u)=\infty$. This constraint improves attack stealthiness and reduces detection probability. The term $\mathcal{S}(.)$ denotes the confidence score that guides adversarial trajectory generation. Score-based attacks primarily seek adversarial trajectories that minimize confidence scores while maintaining proximity within the $\ell_{p}$ norm ball.

\textbf{Decision-Based Optimization}: 
Our decision-based approach operates exclusively on binary decision outputs from adversarial criteria, eliminating gradient dependency. This framework does not require explicit specification of trajectory physical constraints during attack generation. While traditional methods must incorporate kinematic models, velocity limits, and steering constraints, our approach circumvents these requirements by focusing on decision boundary manipulation without modeling underlying physical dynamics. This reduces computational complexity and enhances transferability across different systems. Our optimization objective is formulated as:
\begin{equation}\label{eqn-1} 
  \underset{X_{t}^{*}} {min}\ \mathcal{D} (X_{t} ,X_{t}^{*} )+\zeta(\mathbb{I}(c(X_{t}^{*},\mathbf{F}_{t},n)))
\end{equation}
The distance function $\mathcal{D}$ minimizes perceptible differences between adversarial and original trajectories. The constraint function $\zeta(.)$ performs rejection sampling: $\zeta(u)=0$ if $u$ is true, otherwise $\zeta(u)=\infty$. The indicator function $\mathbb{I}$ describes whether the adversarial criteria is satisfied. Essentially, attackers explore the adversarial region defined by the trajectory prediction model's input domain and adversarial criteria, seeking points guaranteed to be adversarial while remaining as close as possible to the original trajectory.

\textbf{Boundary Walking Algorithm}:
We implement this optimization through a heuristic consisting of two key steps: orthogonal steps ($S_{o}$) and forward steps ($S_{f}$). Orthogonal steps move the adversarial trajectory away from the decision boundary while maintaining distance from the original trajectory, helping escape local optima. Forward steps move the adversarial trajectory closer to the original trajectory.

As shown in Algorithm \ref{alg:dtp_attack}, the algorithm begins by initializing an adversarial trajectory through random sampling near the original trajectory, rejecting non-adversarial samples. After initialization, a forward step brings the adversarial trajectory close to the decision boundary. The algorithm then alternates between orthogonal and forward steps along the adversarial boundary to find the closest adversarial trajectory to the original.

The orthogonal step direction is perpendicular to vector $X_{t}-X^*_{t}$. With sufficiently small step sizes, the orthogonal step approximately satisfies:
\begin{equation}\label{eqn-2}
\mathcal{D} (X^*_{t}+S_{o},X_{t})\approx \mathcal{D}(X_{t} ^*,X_{t})
\end{equation}

Forward steps move in the direction of vector $X_{t}-X^*_{t}$, satisfying:
\begin{equation}\label{eqn-3}
\mathcal{D} (X ^{*(k-1)}_{t}+\eta ^{(k)},X_{t})<\mathcal{D} (X ^{*(k-1)}_{t},X_{t})
\end{equation}

Step sizes are dynamically adjusted according to local boundary geometry. For orthogonal step size $\delta$, we maintain 50\% of orthogonal steps within the adversarial region, adjusting $\delta$ accordingly. Forward step size $\epsilon$ is adjusted to approach the boundary as closely as possible while keeping total step $\eta$ small enough to approximate the boundary as piecewise linear. As distance to the original trajectory decreases, the adversarial boundary becomes flatter and $\epsilon$ gradually decreases. The attack converges when $\epsilon$ reaches zero.

\section{Experiments}
\subsection{Experiment Set-up}

\textbf{Models \& Datasets :}
We conduct experiments on two widely-adopted trajectory prediction benchmarks: nuScenes\cite{caesar2020nuscenes} and Apolloscape\cite{huang2018apolloscape}. Both datasets feature diverse driving scenarios with \SI{2}{\hertz} trajectory sampling. Following standard protocols, we set nuScenes parameters to $L_I = 4$ and $L_O = 12$, and Apolloscape parameters to $L_I = 6$ and $L_O = 6$.

We evaluate two representative trajectory prediction architectures: Trajectron++\cite{salzmann2020trajectron++} and Grip++\cite{li2019grip}. Both models are assessed on each dataset, yielding four base configurations. Additionally, we include Trajectron++(m), which leverages semantic map information available in nuScenes, resulting in five total experimental settings. For multimodal Trajectron++ predictions, we select the mode with highest probability. Each evaluation uses 100 randomly sampled scenarios per dataset.

\textbf{Attack Methods \& Evaluation:}
We compare against two black-box baselines: PSO-based trajectory attack\cite{zhang2022adversarial} and Simple Black-box Attack (SBA)\cite{guo2019simple} from image classification. Given inherent differences in optimization formulations, we establish a unified evaluation framework by first applying our DTP-Attack to determine perturbation bounds, then constraining all baselines within this space to ensure comparable search domains.

We evaluate all methods across two attack objectives with corresponding metrics. For intention misclassification, we measure directional offset distribution (Left, Right, Front, Rear) and Attack Success Rate (ASR) for lane deviation under physical constraints. For prediction accuracy degradation, we assess standard trajectory metrics including Average Displacement Error (ADE), Final Displacement Error (FDE), Miss Rate (MR), Off-Road Rate (ORR), and perturbation magnitude via Mean Squared Error (MSE). Each method operates under a fixed computational budget of 1,000 model queries per scenario.

\textbf{Implementation details:}
Our DTP-Attack method employs a boundary walking algorithm with dynamic step size adjustment. We initialize orthogonal step size $\delta = 1.0$ and forward step size $\epsilon = 0.1$, with adjustment factors of $0.95$ and $0.9$ respectively to maintain algorithmic convergence. The algorithm terminates when $\epsilon <10^{-6}$ or maximum iterations (1,000) are reached.

Adversarial criteria thresholds are configured as follows: For prediction accuracy degradation, $\theta_{err}^{\mathrm{ADE}} = \SI{7.5}{\meter}, \SI{3.5}{\meter}$ and $\theta_{err}^{\mathrm{FDE}} = \SI{17.5}{\meter}, \SI{7.5}{\meter}$ for nuScenes and Apolloscape respectively. For intention misclassification, directional thresholds are set to $\theta_{int}(\mathrm{lateral}) = \SI{2}{\meter}$ and $\theta_{int}(\mathrm{longitudinal}) = \SI{3}{\meter}$ across both datasets. Attack success requires satisfying both adversarial criteria and physical constraints defined by the vehicle dynamics model from\cite{10588723}.

\begin{table*}[t]
\caption{DTP-Attack Performance on Intention Misclassification: Directional Bias Induction and Attack Success Rates}
\centering
\label{Table.result1}
\begin{tabular}{ccccccc}
\hline
\multirow{2}{*}{Model}        & \multirow{2}{*}{Dataset} & Left(m)       & Right(m)      & Front(m)      & Rear(m)       & ASR(\%) \\ \cline{3-7} 
                              &                          & Normal/Attack & Normal/Attack & Normal/Attack & Normal/Attack & Attack  \\ \hline
\multirow{2}{*}{Grip++}       & nuScenes                 & 0.231/2.13    & -0.231/2.21   & -1.02/3.01    & 1.02/3.13     & 81      \\ \cline{2-7} 
                              & Apolloscape              & -0.0131/2.03  & 0.0131/2.18   & -0.0147/3.07  & 0.0147/3.06   & 76      \\ \hline
\multirow{2}{*}{Trajectron++} & nuScenes                 & -0.301/2.03   & 0.301/2.25    & -1.31/3.13    & 1.31/3.21     & 78      \\ \cline{2-7} 
                              & Apolloscape              & 0.0102/2.33   & -0.0102/2.15  & -0.143/3.01   & 0.143/3.11    & 80      \\ \hline
Trajectron++(m)               & nuScenes                 & -0.105/2.01   & 0.105/2.03    & -0.515/3.11   & 0.515/3.05    & 41      \\ \hline
\end{tabular}
\end{table*}

\begin{table*}[t]
\caption{DTP-Attack Performance on Prediction Accuracy Degradation: Error Metrics and Perturbation Analysis}
\centering
\label{Table.result2}
\begin{tabular}{ccccccc}
\hline
\multirow{2}{*}{Model}        & \multirow{2}{*}{Dataset} & ADE(m)        & FDE(m)        & MR(\%)        & ORR(\%)       & MSE(m) \\ \cline{3-7} 
                              &                          & Normal/Attack & Normal/Attack & Normal/Attack & Normal/Attack & Attack \\ \hline
\multirow{2}{*}{Grip++}       & nuScenes                 & 4.34/8.27     & 10.3/18.95    & 16/45         & 3.6/15.3      & 0.21   \\ \cline{2-7} 
                              & Apolloscape              & 1.66/4.68     & 3.18/8.09     & 8/61          & 2.2/28.1      & 0.29   \\ \hline
\multirow{2}{*}{Trajectron++} & nuScenes                 & 3.84/8.15     & 11.2/19.4     & 14/42         & 1.6/12.9      & 0.12   \\ \cline{2-7} 
                              & Apolloscape              & 1.00/4.23     & 2.24/8.71     & 3/55          & 1.5/24.4      & 0.36   \\ \hline
Trajectron++(m)               & nuScenes                 & 1.99/4.05     & 5.37/10.24    & 2/24          & 0.3/5.9       & 0.45   \\ \hline  
\end{tabular}
\end{table*}

\begin{figure*}[!t]
\centering
\includegraphics[width=0.99\textwidth]{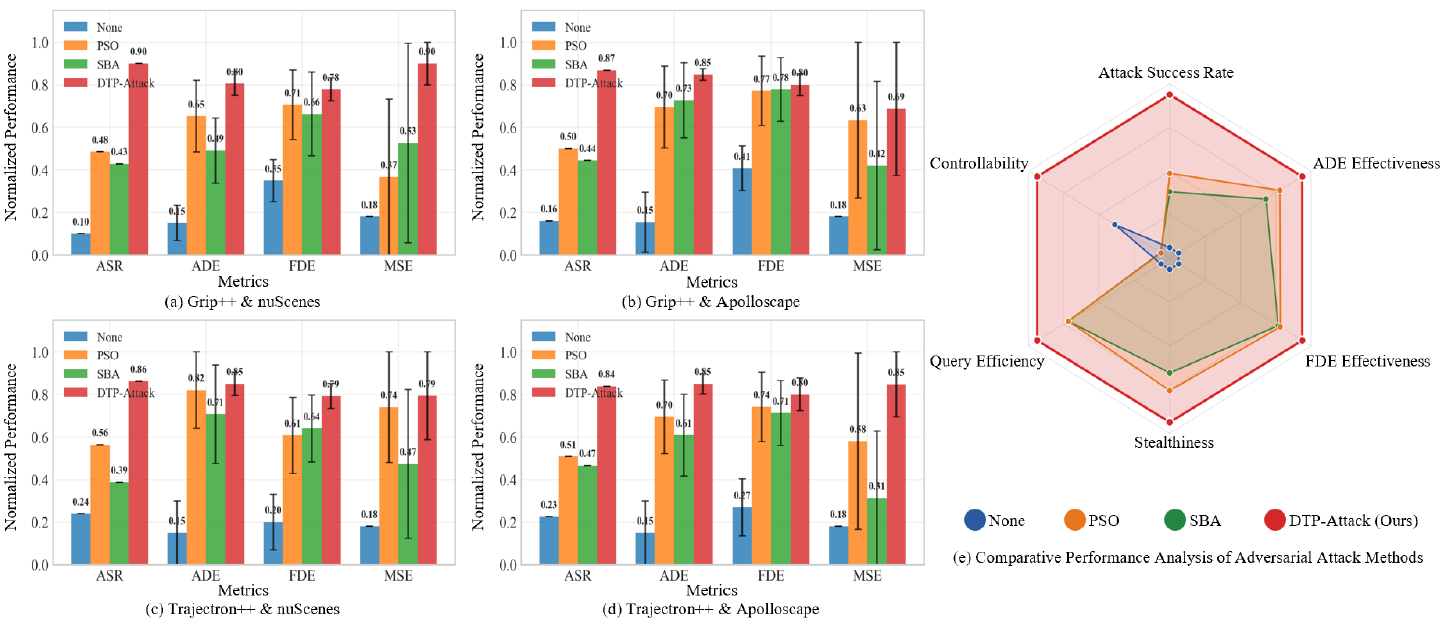}
\caption{Comparative analysis of decision-based vs. score-based adversarial attacks on trajectory prediction models.}
\label{fig:method-comp}
\end{figure*}

\subsection{Main Results}
Our DTP-Attack demonstrates consistent effectiveness across all evaluated models and datasets, successfully achieving both primary attack objectives: intention misclassification and prediction accuracy degradation. The following results reveal fundamental vulnerabilities in current trajectory prediction systems, with potentially severe implications for autonomous vehicle safety.

\textbf{Intention Misclassification:}
Our attacks successfully induce substantial directional biases in trajectory predictions across all configurations, as shown in Table \ref{Table.result1}. Lateral attacks increase prediction deviations from near-zero baselines to over \SI{2}{\meter}, while longitudinal attacks amplify errors from approximately \SI{1}{\meter} to over \SI{3}{\meter}. Importantly, both attack types maintain prediction errors precisely near the adversarial thresholds, demonstrating excellent controllability.

These magnitudes carry significant practical implications. Since urban lane widths typically measure below \SI{4}{\meter}, \SI{2}{\meter} lateral offset causes vehicles to appear as though they will deviate from their designated lanes. Similarly, \SI{3}{\meter} longitudinal offsets create predictions of significant acceleration or deceleration behaviors that fundamentally alter expected vehicle dynamics.

Attack success rates range from $41\%$ to $81\%$ across different model-dataset combinations. Models without semantic map information such as Grip++ and standard Trajectron++ achieve consistently higher success rates than Trajectron++ with map integration, indicating that additional contextual information provides some defensive benefit. However, even this enhanced model remains substantially vulnerable, with success rates exceeding $40\%$. The consistent effectiveness across diverse architectures reveals fundamental vulnerabilities in current trajectory prediction systems, particularly concerning given that our decision-based approach operates without gradient information or detailed model knowledge, conditions that are typical of real-world attack scenarios.

\textbf{Prediction Accuracy Degradation:}
Our attacks cause dramatic accuracy degradation in standard trajectory metrics, as demonstrated in Table \ref{Table.result2}. Average displacement error increases by factors of $1.9\times$ to $4.2\times$, while final displacement error shows similar deterioration with increases ranging from $1.8\times$ to $3.9\times$. These results demonstrate that subtle adversarial perturbations can severely compromise prediction quality across different model architectures.
Miss rates provide particularly striking evidence of attack effectiveness. For Apolloscape, miss rates increase from $8\%$ to $61\%$ for Grip++ and from $3\%$ to $55\%$ for Trajectron++, indicating that most predictions under attack exceed acceptable error thresholds. Off-road rates similarly increase substantially, with predictions frequently placing vehicles in physically impossible locations.

Despite these significant performance degradations, perturbation magnitudes remain remarkably small, with mean squared error between original and adversarial trajectories ranging from \SI{0.12}{\meter} to \SI{0.45}{\meter}. This demonstrates that our method achieves substantial prediction disruption through barely perceptible trajectory modifications that maintain physical plausibility and reduce detection likelihood. These findings highlight the urgent need for robust defense mechanisms in safety-critical autonomous driving applications, where compromised trajectory predictions could lead to catastrophic consequences.

\begin{figure*}[!ht]
\centering
\includegraphics[width=0.87\textwidth]{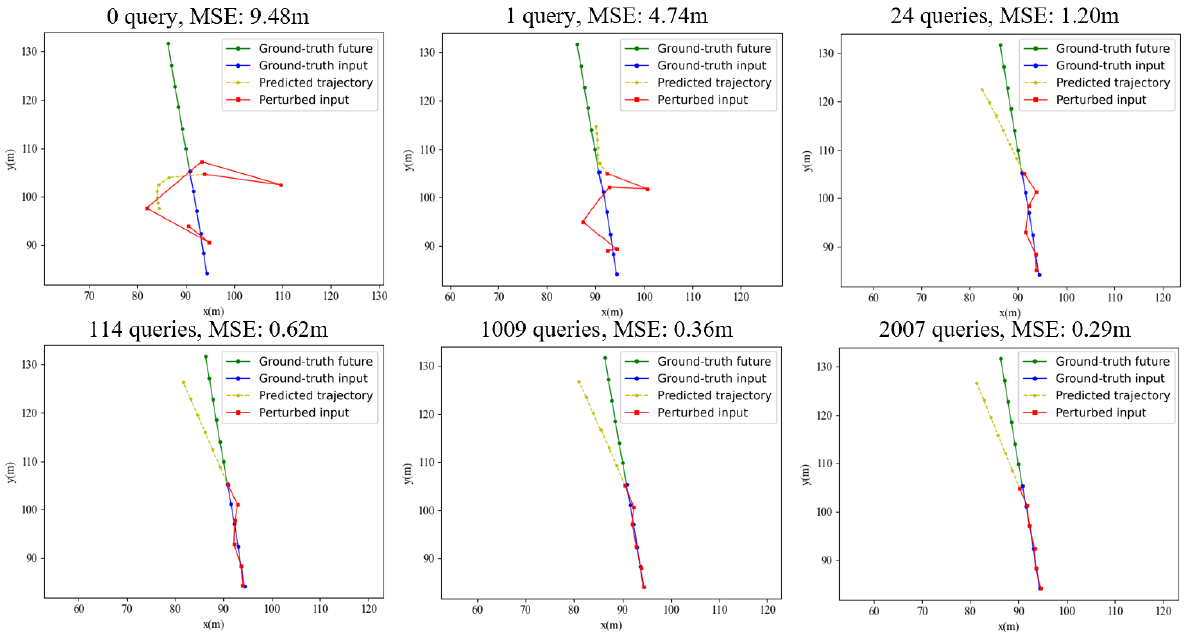}
\caption{Progressive trajectory refinement in DTP-Attack boundary walking algorithm.}
\label{fig:n_calls}
\end{figure*}

\begin{figure}[!t]
\centering
\includegraphics[width=0.87\linewidth]{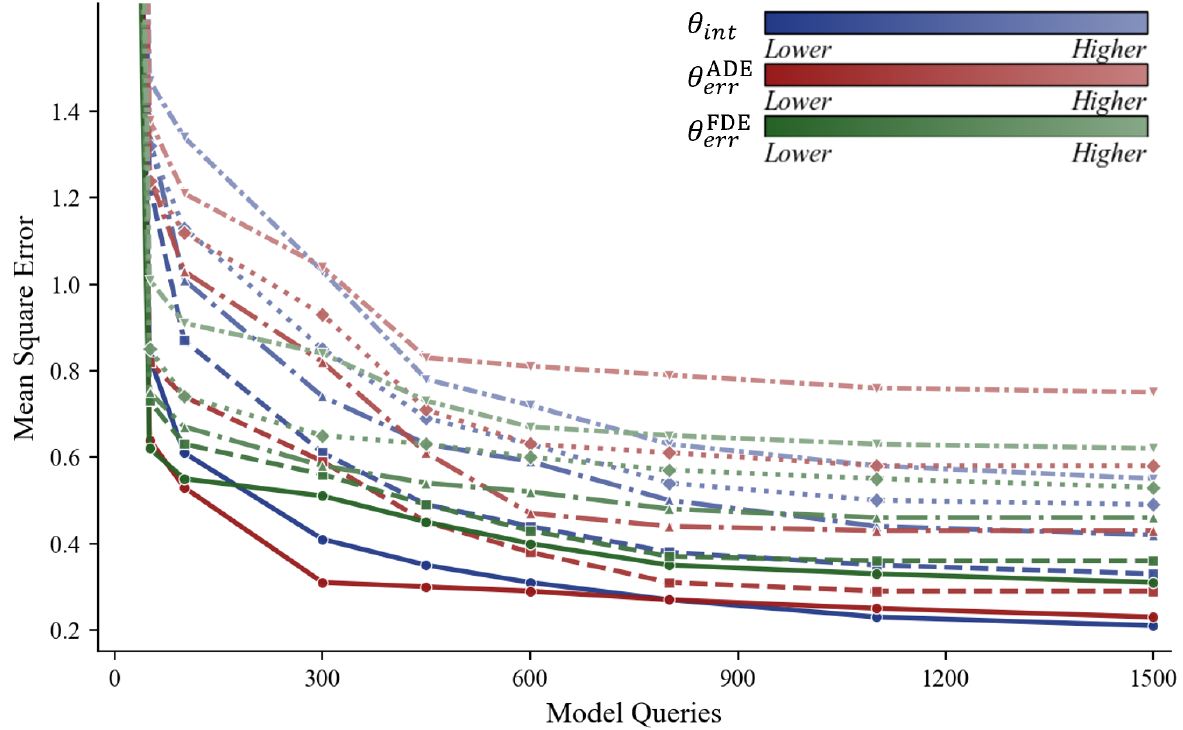}
\caption{Convergence analysis of DTP-Attack under varying adversarial thresholds.}
\label{fig:hyper}
\end{figure}

\subsection{Comparative Analysis}
\textbf{Performance Superiority:}
 Our decision-based approach consistently outperforms existing black-box methods across all normalized evaluation metrics,  as shown in Fig. \ref{fig:method-comp}. The comprehensive comparison reveals substantial improvements in attack effectiveness, with DTP-Attack achieving superior performance on both intention misclassification and prediction accuracy degradation objectives. This consistent advantage holds across diverse model architectures and datasets, demonstrating robust generalizability.
 
A critical advantage lies in attack controllability, evidenced by the exceptionally narrow confidence intervals across $100$ evaluation scenarios. While baseline methods exhibit high variance and unpredictable behavior, DTP-Attack maintains consistently stable output characteristics regardless of scenario initialization or environmental variations. This reliability enables adversaries to execute targeted attacks with predictable outcomes, allowing strategic attack planning where specific prediction failures can be induced with high confidence. The tight error bounds indicate that our boundary walking algorithm reliably converges to near-optimal adversarial trajectories, making it particularly effective for real-world deployment.

\textbf{Methodological Advantages:}
The performance gains stem from fundamentally different constraint handling strategies. Score-based methods incorporate fixed physical constraints to optimize attack stealthiness, explicitly limiting vehicle dynamics and kinematic feasibility. However, these rigid constraints severely restrict the adversarial search space and trap optimization algorithms in local optima. Moreover, fixed constraints cannot adapt across diverse driving scenarios, such as highway maneuvers require different kinematic bounds than urban intersections, limiting method generalizability.
Our decision-based approach circumvents these limitations by avoiding fixed physical constraints entirely. Instead of explicitly enforcing kinematic bounds, we minimize the distance between adversarial and original trajectories, allowing physical plausibility to emerge naturally from proximity preservation. This creates a significantly smoother search landscape without the discontinuous constraint boundaries that cause convergence failures in baseline methods. The resulting optimization surface enables our boundary walking algorithm to navigate complex adversarial regions more effectively, consistently identifying near-optimal solutions and explaining the significantly superior attack success rates.

\subsection{Hyperparameter Analysis and Case Study}
Fig. \ref{fig:n_calls} demonstrates iterative attack evolution from initial random sampling (MSE: \SI{9.48}{\meter}) to optimized results (MSE: \SI{0.29}{\meter} after $2007$ queries) under Trajectron++\&Apolloscape. The boundary walking algorithm achieves rapid initial improvement, with MSE dropping to \SI{4.74}{\meter} after one query and \SI{1.20}{\meter} after $24$ queries, representing $50\%$ and $87\%$ reductions respectively.

The optimized attack generates prediction errors exceeding physical lane boundaries. In a scenario where an autonomous vehicle drives in the left adjacent lane, false left-turn predictions would trigger collision avoidance responses including emergency braking. The minimal perturbation magnitude demonstrates the feasibility of generating imperceptible attacks that influence trajectory prediction outputs.

We evaluate attack performance across varying threshold difficulties: $\theta_{int}\in \left \{ 2,3,4,5,6 \right \}$, $\theta_{err}^{\mathrm{ADE}}\in \left \{ 4,5,6,7,8 \right \}$, and $\theta_{err}^{\mathrm{FDE}}\in \left \{ 17,18,19,20,21 \right \}$. Fig. \ref{fig:hyper} shows convergence exhibits a consistent two phase pattern: rapid MSE reduction within $300$ queries followed by asymptotic stabilization.
Lower adversarial thresholds produce smaller final MSE values, confirming that relaxed attack criteria enable closer approximation to original trajectories. This validates our optimization framework and demonstrates the trade-off between attack success and perturbation stealthiness.

\section{Conclusions}
We present DTP-Attack, the decision-based black-box adversarial attack framework for trajectory prediction systems in autonomous driving. Unlike existing methods requiring gradient information or model internals, our approach operates exclusively on binary decision outputs, making it applicable to real-world scenarios. Our novel boundary walking algorithm navigates adversarial regions without fixed physical constraints, naturally maintaining trajectory realism through proximity preservation. Extensive evaluation on nuScenes and Apolloscape datasets across state-of-the-art models demonstrates superior performance, achieving $41-81\%$ attack success rates for intention misclassification with perturbations below \SI{0.45}{\meter}, and increasing prediction errors by $1.9-4.2\times$ for accuracy degradation. These minimal perturbations can induce \SI{2}{\meter} lateral and \SI{3}{\meter} longitudinal prediction deviations, potentially causing autonomous vehicles to misinterpret critical driving maneuvers. Our method consistently outperforms existing black-box approaches while revealing fundamental vulnerabilities in current trajectory prediction systems. These findings highlight urgent needs for robust defenses in safety-critical autonomous driving applications. Future work should focus on developing defensive mechanisms that can detect and mitigate such adversarial attacks while maintaining prediction accuracy under normal conditions.

\bibliographystyle{unsrt}
\bibliography{references}

\end{document}